# Wearable environmental sensing to forecast how legged systems will interact with upcoming terrain

Michael D. Murray[1], James Tung[1], Richard W. Nuckols[1,2*]

[1] Systems Design Engineering, University of Waterloo; Waterloo, ON, Canada;

[2] Department of Mechanical and Industrial Engineering; University of Massachusetts Lowell, Lowell, MA, USA

*Correspondence to: richard_nuckols@uml.edu

**Abstract:** Computer-vision (CV) has been used for environmental classification during gait and is often used to inform control in assistive systems; however, the ability to predict how the foot will contact a changing environment is underexplored. We evaluated the feasibility of forecasting the anterior-posterior (AP) foot center-of-pressure (COP) and time-of-impact (TOI) prior to foot-strike on a level-ground to stair-ascent transition. Eight subjects wore an RGB-D camera on their right shank and instrumented insoles while performing the task of stepping onto the stairs. We trained a CNN-RNN to forecast the COP and TOI continuously within a 250ms window prior to foot-strike, termed the forecast horizon (FH). The COP mean-absolute-error (MAE) at 150, 100, and 50ms FH was 29.42mm, 26.82, and 23.72mm respectively. The TOI MAE was 21.14, 20.08, and 17.73ms for 150, 100, and 50ms respectively. While torso velocity had no effect on the error in either task, faster toe-swing speeds prior to foot-strike were found to improve the prediction accuracy in the COP case, however, was insignificant in the TOI case. Further, more anterior foot-strikes were found to reduce COP prediction accuracy but did not affect the TOI prediction accuracy. We also found that our lightweight model was capable at running at 60 FPS on either a consumer grade laptop or an edge computing device. This study demonstrates that forecasting COP and TOI from visual data was feasible using a lightweight model, which may have important implications for anticipatory control in assistive systems.

## INTRODUCTION

Humans adapt their gait when navigating complex and dynamic terrains. On less dynamic terrains such as walking on level ground, gait generally follows regular cyclic patterns in which characteristic lower-limb joint dynamics can be modeled and predicted (*1*). However, these predictable patterns break down during transitions between terrains. For instance, the precise musculoskeletal torque required to efficiently transition from a level-ground state to a stair-ascent state depends on the foot placement on the stair and the relative location of the foot center of pressure (COP) and the ankle joint center upon environmental contact (*2*).

For seamless and stable control, the visual system and muscles must operate in a coordinated feedforward fashion, proactively anticipating obstacles or terrain changes and initiating adaptive torques before physical contact occurs (*3–5*). In cases where this synchrony is impacted, such as due to visual impairments, gait adaptability is often compromised which can lead to an increased risk of instability or falls (*6, 7*). These challenges highlight the importance of vision in anticipatory control mechanisms, and, in the context of legged locomotion in robotics, provides insight on factors important for the design of anticipatory mechanisms.

Given the importance of environmental awareness, computer vision has been applied in robotics (*8–12*), ground vehicles (*13–15*), and assistive devices for individuals with vision impairments (*16–20*). This has also been recently applied to lower-limb assistive robotics where vision-based terrain classification has enabled these systems to anticipate environment changes and switch discrete assistance profiles accordingly (i.e. finite-state machines (FSM)) (*21–24*). FSMs for lower limb-assistive systems to often have states for level-ground, stair ascent, stair descent, and sometimes transition states (*21, 23–25*). While FSM-based systems support general planning, they do not provide sufficient information regarding the gait transition, such as foot placement and impact timing, that is required for joint level control strategies.

First, while these classification systems have generally performed well when the visual input contains only one terrain class, achieving accuracies of 73%, 95%, 91.67%, and 93% across 4 studies respectively (*21, 23–25*), they performed much more poorly when multiple terrains were visible simultaneously such as the visual field including both level ground and stair terrains. Hence, the same studies achieved accuracies of only 49.1%, 33%, 47.6%, and 59.3% respectively when visual inputs resembled level-ground to stair ascent transitions (*21, 23–25*). This aligns with broader challenges in labelling transitions, where subjectivity in defining terrain boundaries complicates model training (*26*).

Second, current FSM based systems in literature do not handle more specific but important biomechanical details, such as foot-strike strategy (e.g., rear-foot-first vs fore-foot-first), which can influence ankle moment required for effective gait (*2*). For example, when stepping on a stair during stair ascent, fore-foot-strikes (COP closer to the fore-foot) result in a larger plantar-flexion moment ($0.46 \pm 0.16$ Nm kg$^{-1}$ (RMS $\pm$ SD)) during the weight acceptance phase compared to rear-foot-strikes ($0.09 \pm 0.05$ Nm kg$^{-1}$) due to longer moment arms in the former case (*2*). Furthermore, in the situation where the COP and ground reaction force are posterior to the ankle center of rotation, the ankle moment will switch signs (*27*). Hence, if an FSM system applies the same torque for a task regardless of foot placement, it may apply inappropriate assistance when the user lands rear-foot first rather than fore-foot first. This could result in instability and increased metabolic costs as the user may have to resist or stiffen their joint to respond to inappropriate or unexpected assistance.

Humans naturally adapt their gait in anticipation of terrain changes by modulating their muscle activation and joint mechanics based on visual cues (*28–31*). To emulate this anticipatory control within assistive systems, we must extend the current research to not only understand *what* terrain lies ahead, but *where* and *when* the foot will make contact. In this work, we propose an approach to forecast two key parameters during stair ascent transitions. First, the anterior-posterior (AP) foot center of pressure (COP) upon impact on the stair will indicate *where* the impact will occur, influencing required torques for stability and propulsion. Second, the time of impact (TOI) can indicate *when* the foot impact occurs and could be used in appropriate timing of the applied torque or potentially other assistance strategies in real-time. Both quantities could provide invaluable insight into the spatial and temporal dynamics of foot placement, critical for informing exoskeleton, assistive device, or legged robot control strategies. Prior studies have estimated COP *trajectory* during gait based on visual odometry (*32*) or insole/IMU based models (*33*). These studies, however, only estimate COP *trajectory* after the foot has already contacted the surface. Additionally, although other studies have explored TOI estimation, existing methods have been generally intended for noncausal biomechanical analyses and rely on 3D motion capture systems (*34*, *35*). In contrast, predictive, pre-contact information regarding where and when foot–surface contact will occur may be valuable both to the user and for informing assistive device control.

Most studies to date which perform vision-based terrain classification use machine learning (ML), often fine-tuning larger pre-trained architectures such as EfficientNet (*21*), YOLOv8 (*24*), and ResNet-18 (*25*). These models were often trained on either custom or public datasets such as ExoNet (*26*). Despite the large scale of ExoNet, it consists of frames labelled with their associated class at a frequency of only 5 frames-per-second (FPS). These sparse labels may be sufficient for high-level terrain classification used in FSM based control, where higher model complexity (and thus latency) was acceptable and more frequent predictions were redundant. In contrast, our task of estimating COP and TOI requires capturing the dynamics and precise timing of foot swing and placement, necessitating a higher temporal resolution (estimated at 30-60 FPS). Further, if deployed in an assistive system, predictions must occur frequently enough to compensate for the electromechanical delay of the system, motivating a lightweight ML architecture capable of real-time inference.

In this work, we developed a data-driven approach to train a machine learning framework forecasting the impact COP (*where*) and TOI (*when*) during swing phase before the foot-stair contact occurs. Our approach used a combined convolutional and recurrent neural network (CNN-RNN) architecture (*36*, *37*) to extract and process lower limb and environmental features from data captured from a stereoscopic camera at 60 FPS. This work highlights potential for exoskeleton control, but the broader framework may extend to other applications, including mobility aids for the visually impaired and legged robots.

## RESULTS
## COP Forecast Results

To evaluate the viability of vision-based COP forecasting in dynamic and complex conditions, we developed a vision-based system incorporating an RGB-D camera (OAK-D Wide, Luxonis, Denver, CO) and an ML model. The ML model estimated the future impact foot-strike COP location (termed forecasted COP (fCOP)) at the first stair of a stair ascent transition from level ground based on the visual input from the camera (Fig. 1). When referring to the estimation of future events, the term 'forecasting' is often used instead of 'prediction' to emphasize the temporal aspect of the estimation. The stereoscopic RGB-D camera, mounted on the subjects' shank and oriented toward the foot, recorded images during gait. Instrumented insoles (Moticon,

Moticon AG, Munich, Germany) provided ground-truth AP COP data. To introduce gait variability during the stair approach and transition, subjects were instructed to approach the stair at three different speeds (slow = 1.0 m s⁻¹, medium = 1.25 m s⁻¹, fast = 1.5 m s⁻¹) and to employ three different foot-strike strategies (i.e., rear-foot, mi—foot, or fore-foot) upon stair contact  (see Supplementary Fig. S1 for approach velocity and foot-strike strategy distributions). The recorded camera images of the participants final ipsilateral swing phase and contact at the first stair were used to train a machine learning model that generated fCOP using a constant lookback window of 250ms. We collected data from 8 healthy, young participants [n = 8, 2 females and 6 males; age = 22.8 ± 1.9 years (means ± SD); mass = 66.6 ± 5.7 kg; height = 172.5 ± 7.1 cm], each performing approximately 90 trials of the level ground to stair ascent transition. Participants wore their own shoes for the study.

Our CNN-RNN model received grayscale stereoscopic frames, down sampled to a 50x25 resolution, and stacked along the channel dimension, as input into the CNN layers. The extracted features from the CNN were then input into an RNN for temporal processing across frames and for final prediction. We trained the model to estimate the impact fCOP at discrete timepoints defined by the camera's frame rate of 60 FPS (i.e. forecast every 16.67 ms). The time difference between the time of forecast and the actual time of contact was termed the forecast horizon (FH). To form predictions on the dataset, we performed leave-one-out-cross-validation (LOOCV).

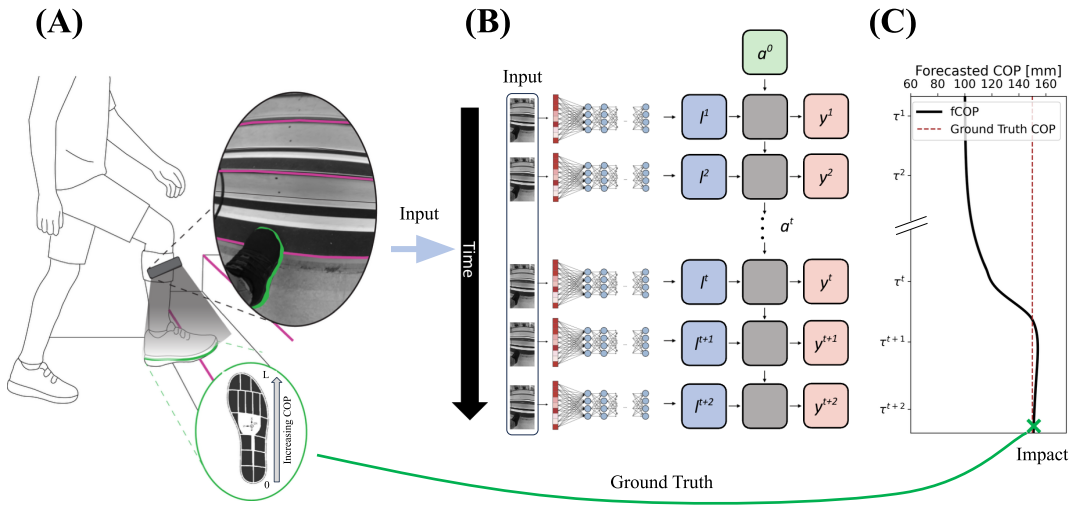

**Fig. 1. Prediction process demonstrating the COP task. (A)** *The subject was equipped with a camera attached to shank and a sensorized insole in their shoe. Insole ground truth values vary from 0mm at the rear-foot to the length of the insole (L) at the fore-foot. Colors are meant for clarity of figure and were not generated by model.* **(B)** *The machine learning model formulates a prediction at each timestep (frame) based on the current input frame and feedback based on past frames. l represents the latent feature vector extracted from the CNN, a represents the hidden state propagated through time for the RNN, and y represents the output fCOP at each timestep* **(C)** *A theoretical fCOP prediction curve which approaches the ground truth COP as the foot nears foot-strike.*

To evaluate the effectiveness of the model, we computed the error between the actual COP location at foot-strike and the forecasted COP location every 16.67 milliseconds (ms) from 16.67 milliseconds to 250 milliseconds prior to foot-strike. The group mean-absolute-error (MAE) in fCOP decreased relatively linearly as the FH decreased ($R^2 = 0.98, p < 0.001$) (Fig. 2A). The MAE was 29.42, 26.82, and 23.72 mm for forecast horizons at 150 ms, 100 ms, and 50 ms, respectively. These errors represented about 11.2, 10.2, and 9.0% of the total foot length of the average North American (NA) foot (263.2mm (*38*)), respectively. The subject level results were similar to that of the group, demonstrating linearity in performance decay (Subject 2 shown in Fig. 2B, others can be found in Supplementary Fig. S2).

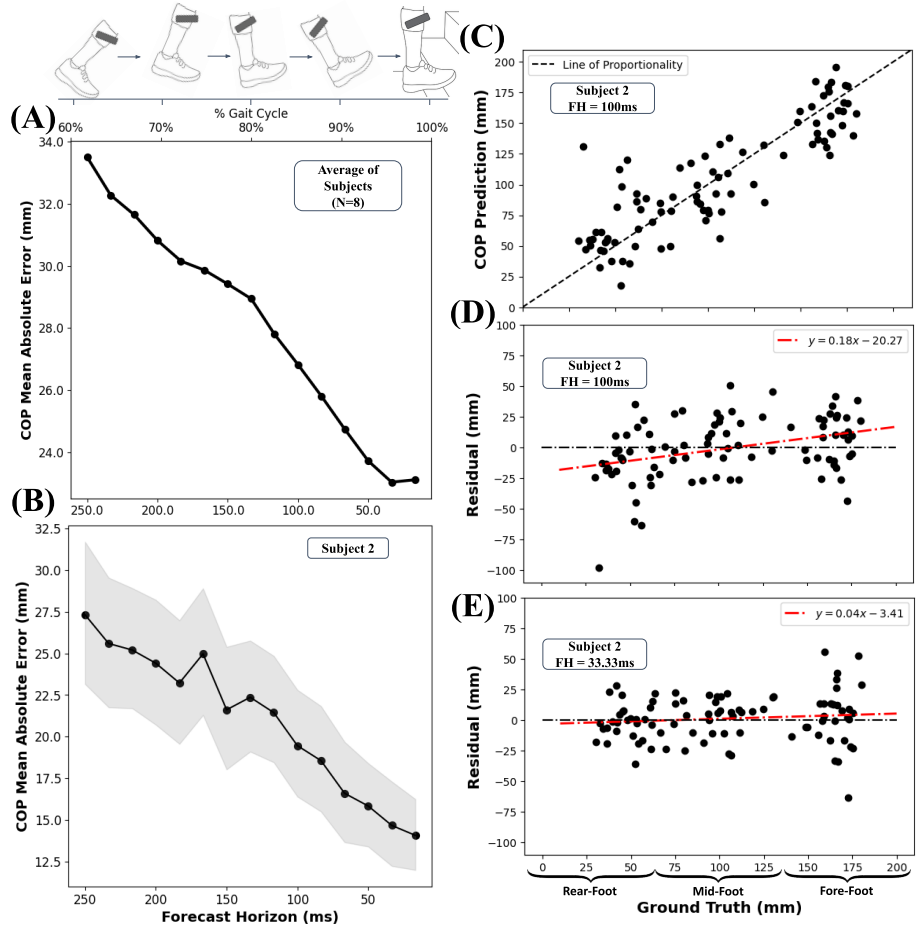

***Fig. 2. fCOP prediction results as a function of FH and ground truth COP. (A)*** *The group mean MAE-FH curve with depiction of the ipsilateral foot approaching the stair through the gait cycle (assuming 1.0 second stride). Circular markers represent each FH (1/60s period)* ***(B)*** *MAE-FH curve for Subject 2 with 95% bootstrap (B=10,000) confidence error shading.* ***(C)*** *Scatter plot of prediction vs ground truth COP for Subject 2 at a 100.0 ms FH.* ***(D)*** *Residual plot with linear regression for Subject 2 at a FH of 100ms.* ***(E)*** *Residual plot with linear regression for Subject 2 at a FH of 33.33ms.*

To better understand the effect of foot strike location on error and provide insight beyond aggregate measures such as MAE, we analyzed how the prediction error distributes at specific FHs. As an example, the relationship between fCOP and ground truth COP was relatively linear at a FH of 100ms ($R^2 = 0.75, p < 0.001$), with an MAE of 19.46mm (95% CI [16.41, 22.81]) for *Subject 2* (SJ2, Fig. 2C). All subjects are shown in Supplementary Fig. S3.

We used error residuals, $r_k = y_k - \widehat{y_k}$ (the difference between the actual ($y_k$) and predicted ($\widehat{y_k}$) COP) to assess how prediction accuracy changed along the length of the foot at impact, and whether predictions erred towards the midfoot or the extremes. Negative residuals ($r_k < 0$) indicate predictions that were greater than the ground truth (overprediction) and positive residuals ($r_k > 0$) indicate predictions that were less than the ground truth (underprediction). To assess the general trend, we modelled the residuals as a function of the ground truth in a linear regression ($r \sim y$).

For SJ2, linear regression slope coefficients suggested a "conservative", slightly positive, tendency of prediction towards the mean of the data (towards the midfoot), with $\beta_1 = 0.18$ and $\beta_1 = 0.04$ at a FH of 100.0ms (Fig. 2D) and 33.33ms respectively (Fig. 2E). Similar phenomena were seen for all subjects (Supplementary Fig. S4). Generally, these slope coefficients were correlated with increasing FH (Subject average Pearson's r = 0.9143 (SD=0.0910)), meaning that as forecasts were formulated earlier, the model took a more conservative estimate.

Further, at the FH level, the residuals were not systematically biased towards over- or under-predicting the COP location along the foot (*SJ2* in Fig. 3A, others in Supplementary Fig. S5). For example, the median residual at a FH of 100.0ms for SJ2 was -2.2mm. Meanwhile, the variance in the prediction generally increased with FH, which led to increased error overall (see Supplementary Fig. S5, S6). Further, at any given FH, there was some inter-subject variability in the residual distributions (subject min IQR = 29.53mm, subject max IQR=53.84mm at a FH of 100ms), (Fig. 3B, Supplementary Fig. S5)).

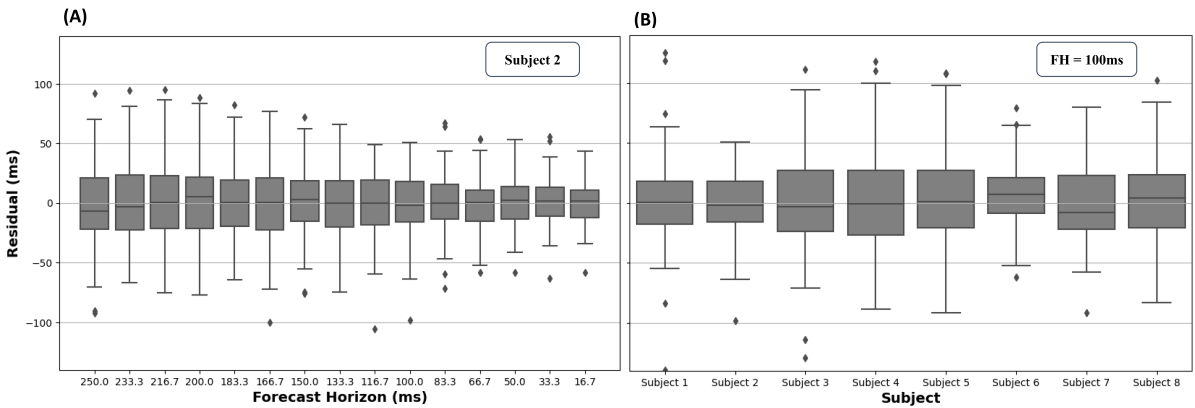

**Fig. 3. Boxplots of residuals for the COP task.** *(A) Boxplots of residuals over FH for Subject 2. (B) Boxplots of residuals for all subjects at a fixed forecast horizon of 100.0 ms.*

**TOI Forecast Results**

We used the same images used in the COP task to train a second causal model that predicted the foot's TOI (fTOI) during the first step of the stair-ascent transition. To evaluate the effectiveness of the model, we calculated the mean-absolute-error between fTOI and the ground truth TOI at each FH. When training the model, only the images at the FH and prior (up to a constant lookback time of 250ms) were provided to the model to ensure that causality was maintained.

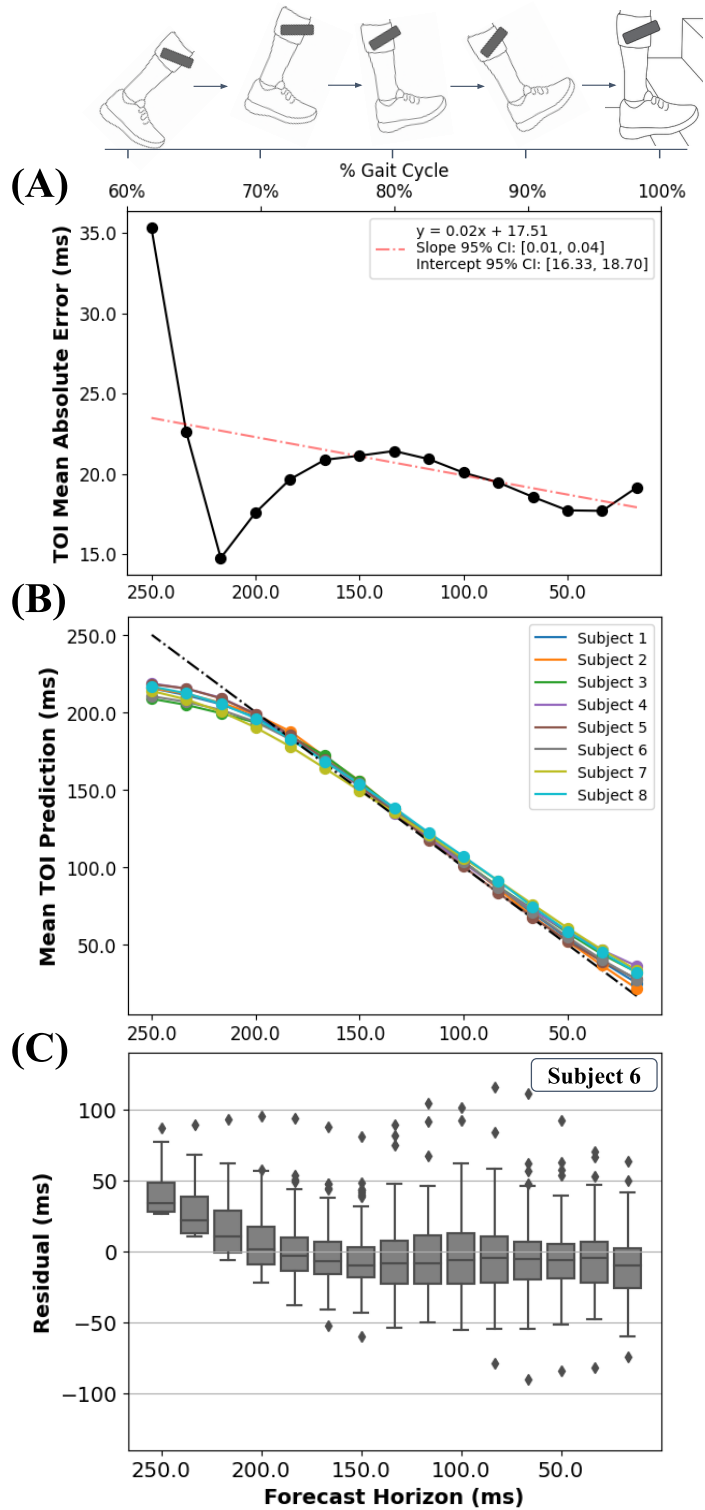

**Fig. 4. fTOI prediction results as a function of FH. (A)** *The group mean MAE-FH curve with depiction of the ipsilateral foot approaching the stair through the gait cycle (assuming 1.0 second stride). Circular markers represent each FH (1/60s period).* **(B)** *Mean fTOI curves for all subjects compared to the idealized prediction line.* **(C)** *Boxplots of residuals for each FH (1/60s period) for SJ6.*

Due to the nonlinear behaviour at longer FH, we performed a piecewise analysis with the split at a FH of 166.67ms. From a FH of 166.67ms to 16.67ms, the fTOI MAE decreased relatively linearly from 20.88ms to 19.16ms ($R^2 = 0.74, p < 0.005$) (Fig. 4A). While most subjects experienced a negative slope in this region, some subjects had negligible slopes (see Supplementary Fig. S7). At a FH of 216.67ms, the MAE decreased to 14.76ms before it increased sharply afterwards for longer FH (Fig. 4A). This effect resulted from when the model began to default towards prediction of constant values when FH increased, which generally started around a FH of 166.67ms (see Supplementary Fig. S8). The precise value at which the model regressed towards was subject dependent (Fig. 4B, subject-level; min = 208.68ms, max = 218.60ms).

Subjects showed relatively low inter-subject variability in their mean prediction curves, which were generally unbiased except at the extremes (Fig. 4B). At low FHs, predictions tended to be higher than the ground truth (i.e., predicting the impact will occur later than it actually occurred), and for high FHs, predictions tended to be lower (Fig. 4B, Fig. 4C, Supplementary Fig. S9 for all subjects).

While low to mid FHs (i.e., 100.0ms) had residual distributions that appeared relatively normal (Fig. 4C for *SJ6*, Supplementary S10 for all subjects, Fig. 5), residuals became right skewed for larger FH (Fig. 4C for SJ6, Supplementary S11 for all subjects). Further, at any given FH, there was some inter-subject variability in the residual distributions (subject min IQR = 26.82ms, subject max IQR= 49.13ms at a FH of 100ms, (Fig. 5, Supplementary Fig. S9)).

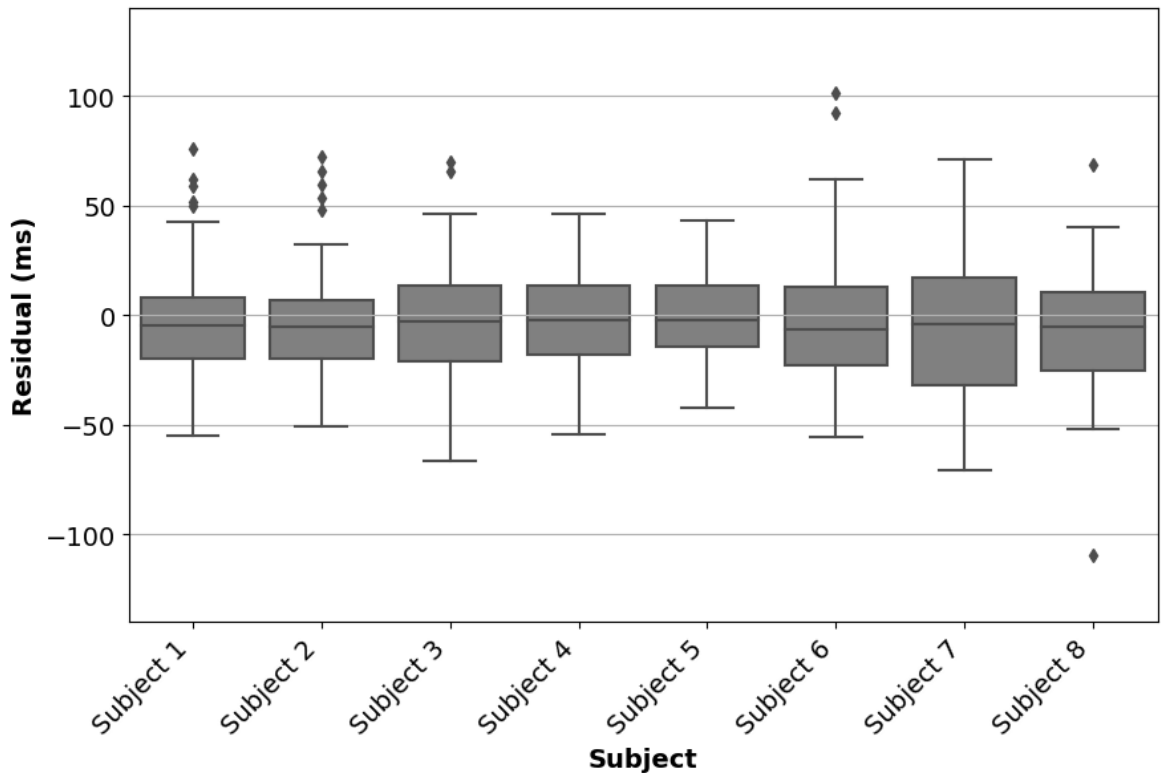

*Fig. 5. Boxplots of residuals for all subjects for the TOI task at a fixed forecast horizon of 100.0 ms.*

**COP Error Mixed Effect Linear Modelling**

To investigate the factors that influence the COP prediction error, we fit a linear mixed effects model (LMM) with forecast horizon (FH), torso velocity, toe velocity, and COP ground truth as predictors. We used the cube-root transformed absolute COP error as the response variable to satisfy homoscedasticity, and appropriately back transformed the results for interpretation (see Materials and Methods). Subject-level random effects were included for the intercept and FH slope.

The FH, toe velocity, and COP ground truth were significant model predictors of the transformed absolute COP error ($p<0.05$, Table 1). However, torso velocity was not a significant predictor and was correlated to toe velocity (correlation matrix in Supplementary Table S2), and hence, it was removed upon a subsequent model fit. In the re-fit model, FH, toe velocity, and COP ground truth remained statistically significant ($p<0.05$, Table 1).

**Table 1. Fixed effects from COP mixed effect linear modelling.**

| Predictor | Domain | Expected Abs Err. Increase Across Domain† (mm/[unit]) | p-value |
|---|---|---|---|
| Intercept | N/A | 26.74 **\*27.94** | <0.001 **\*<0.001** |
| Forecast Horizon (ms) | [16.67, 250] | 10.35 **\*10.35** | <0.001 **\*<0.001** |
| Torso Velocity (mm s$^{-1}$) | [1000, 1500] | 1.29 | >0.05 |
| Toe Velocity (mm s$^{-1}$) | [3000, 5000] | -2.66 **\*-1.60** | <0.01 **\*<0.05** |
| Ground Truth COP (mm) | [53.34, 172.75] ‡ | 3.028 **\*3.12** | <0.001 **\*<0.001** |

\*Represents the estimate after removing torso velocity.
† The domain specifies the predictor end-points at which the mixed effects model was used to predict the error, with all other predictors held at their mean. The difference in predicted error between these two points represents the values in this column.
‡ Domains which use the 25th and 75th percentiles as domain endpoints

We assessed the LMM model coefficients to understand the impact of each relevant predictor on error. We first investigated the fixed-intercept ($B_0 = 27.94$mm) which represents the absolute error when all predictors were at their mean (due to the standardization of variables). Hence, it was the estimate of the average absolute error for when the FH was 133.33ms, toe velocity was 4154.72 mm s$^{-1}$, and COP ground truth was 113.25mm (representing the midfoot). Then, we evaluated slope coefficients to examine the strength of the relationship between the predictors and the absolute COP error. Our model coefficients revealed a positive association between FH and absolute error. Based on the coefficient, the expected average increase in absolute error from a FH of 16.67ms to a FH of 250.00ms was 10.35mm, which corresponds to 3.93% of the average NA foot (*38*). In contrast, we found an inverse relationship between toe velocity and absolute error. In our study, toe velocities mostly ranged between 3000 and 5000 mm s$^{-1}$. Given this, we would expect a decrease in error of 1.60mm between the slower to the faster speeds (0.61% of the average NA foot length). We also observed that the COP ground truth (i.e., the true foot strike location) exhibited a positive relationship with error, where foot placements that were more anterior (towards the fore-foot) were associated with larger errors. For example, our analysis demonstrated that COP ground truth locations at the 75th percentile (172.75 mm) tend to have errors about 3.1238mm greater than those at the 25th percentile (53.34 mm).

The random intercepts varied across subjects with a standard deviation of 4.52mm (Table 2), meaning that with all other predictors at their mean values, we expect differences of ~4.5mm between subjects. The contribution to between-subject variability due to FH slopes was similar, as the error differential due to the slope between the FHs of 16.67ms and 250.00ms varied with a standard deviation of 4.66mm across subjects (Table 2).

**Table 2. Standard Deviation of random effects from COP and TOI mixed effect linear modelling.** The random intercept has the same unit as the representative quantity (mm for COP, ms for TOI). The random slope for the COP task has units of mm/ms, while the TOI task is unitless (ms/ms).

| | Standard Deviation | |
|---|---|---|
| **Quantity** | **COP (mm)** | **TOI (ms)** |
| Random Intercept | 4.50 | 2.55 |
| ([unit]) | **\*4.52** | **\*2.51** |
| Random Slope of Error | 4.64 | 1.87 |
| w.r.t Forecast Horizon | **\*4.66** | **\*1.87** |
| ([unit]/ms) | | |

**\*** Represents the estimate after removing insignificant effects.

## TOI Error Mixed Effect Linear Modelling

We performed analogous LMM modelling for the TOI task with respect to the (cube-root transformed) absolute TOI error. Since the MAE curve with respect to forecast horizon (FH) was non-linear, we studied a restricted linear domain up to a FH of 166.67ms.

FH was the only significant predictor of absolute TOI error ($p<0.05$, Table 3). Contrastingly, torso velocity, toe velocity, and COP ground truth were not significant predictors ($p>0.05$) in this analysis. We then removed these insignificant effects and refit the model.

We also analyzed the model coefficients for the TOI analysis. The fixed-effect intercept, representing the error for when all predictors were at their mean (FH was 91.67ms, toe velocity was 4154.72mm s$^{-1}$, and COP was 113.25mm), was $B_0=19.77$ms (Table 3). Similar to the COP analysis, we evaluated the slope coefficients to further investigate how the FH (the only significant effect) related to the error. In this case, predictions formed earlier before the impact (higher FH) resulted in higher errors compared to those predictions formed closer to the impact (smaller FH). For example, predictions formed at a FH of 166.67ms produced errors about 3.079ms greater than those at a FH of 16.67ms (Table 3). This error corresponds to about 18.47% of the time between two consecutive frames (60 frames – 16.67ms period).

**Table 3. Fixed effects from TOI mixed effect linear modelling.**

| Predictor | Domain | Expected Abs Err. Increase Across Domain[†] (mm/[unit]) | p-value |
|---|---|---|---|
| (Intercept) | N/A | 19.77 **\*19.77** | <0.001 **\*<0.001** |
| Forecast Horizon (ms) | [16.67, 166.67] | 3.078 **\*3.079** | <0.05 **\*<0.05** |
| Torso Velocity (mm s$^{-1}$) | [1000, 1500] | -0.73 | >0.05 |
| Toe Velocity (mm s$^{-1}$) | [3000, 5000] | -0.047 | >0.05 |
| Ground Truth COP (mm) | [53.34, 172.75] [‡] | -0.69 | >0.05 |

\* Represents the estimate after removing insignificant effects.
[†] The domain specifies the predictor end-points at which the mixed effects model is used to predict the error, with all other predictors held at their mean. The difference in predicted error between these two points represents the values in this column.
[‡] Domains which use the 25th and 75th percentiles as domain endpoints

Random intercepts varied across subjects with a standard deviation of 2.51ms (Table 2), meaning that, with all other predictors at their mean values, we might expect differences in error between subjects of ~2.5ms (within the FH range of 16.67ms to 166.67ms). Further, the between-subject variability due to FH slopes was 1.87ms (Table 2).

**Live Simulation**

To quantify the system's runtime performance, we assessed the effective frame rate of the camera when running both the COP and TOI models simultaneously and continuously through a MacBook Air (Apple M2 chip, 16GB unified memory) and a Jetson Orin Nano (8 GB LPDDR5, 6-core ARM Cortex-A78AE CPU, 1024-core Ampere GPU). Frame rate tests (m=3 trials) over 2-minute intervals which achieved a mean FPS of 59.954 FPS (SD = 0.001 FPS) and 59.951 FPS (SD=0.0003 FPS) for the MacBook and Jetson respectively.

**DISCUSSION**

This study demonstrates the feasibility of forecasting both the anterior-posterior center of pressure (COP) and the time of impact (TOI) on a level-ground to stair-ascent transition, using a leg-mounted RGB-D camera and a lightweight machine learning model. Inspired by evidence that human muscle activation and joint mechanics adapt before foot strike (*28–31*), our work aims to achieve similar anticipatory and proactive strategies for wearable technologies. This could have applications in exoskeletons or other lower-limb assistive robotics across various contexts, including rehabilitation (*39–41*), daily mobility assistance (*32*, *41*, *42*), and in physically demanding occupations (*41*, *43*).

Previous literature has attempted to address the challenge of terrain adaptation from the perspective of terrain classification and FSM state switching. However, such solutions tend to have drastically reduced accuracies upon transitions (i.e., from level-ground to stair-ascent) (*21*, *23–25*). Furthermore, they are agnostic to how local biomechanical details influence the torques necessary for stable and efficient gait transitions. For instance, when transitioning from level ground to stair ascent, the musculoskeletal torque required depends on the foot-placement strategy, which has been often neglected in the literature (*2, 21–24*). The lack of fine-grained context in current systems could be seen analogous to a self-driving car that is capable of detecting an obstacle in its path, however, lacks information about where the car is relative to the obstacle and when the interaction between the car and the obstacle may occur. In our study we focus on fine-grained details of these transition states, specifically predicting *where* (fCOP) and *when* (fTOI) the human foot will make contact with a stair during the transition from level ground into stair ascent.

We first forecasted COP and evaluated the error at FHs between 16.67ms and 250.00ms. Expectantly, absolute COP prediction error increased linearly with larger FH. This was consistent with the expectation that uncertainty will grow with increasing FH as the model must extrapolate its current state further into the future. At larger FHs, the model exhibited a conservative bias towards midfoot predictions, which may reduce the likelihood of extreme misclassifications when under uncertainty. This could be advantageous in applications such as ankle exoskeletons (*44–46*), where largely incorrect predictions could result in large and inappropriate torque application (i.e., applying dorsiflexion when plantarflexion was necessary). Conservative estimates would diminish the consequences of such incorrect predictions.

Other studies have performed COP estimation, however in a fundamentally different way. For example, one study estimated the COP trajectory during stance after foot-strike using visual odometry. Their method achieved an overall COP root-mean-square-error (RMSE) of 20-26mm and 28mm for level ground and stair ascent environments respectively (*32*). Another study, which also estimated the COP trajectory during stair ascent, achieved an RMSE of 17.94mm along the AP axis by fusing post foot-strike pressure insole data with leg-mounted inertial measurement units (IMUs) (*33*). While we predict the *impact* center of pressure *before* foot-strike, these studies predict the COP *trajectory* only *after* foot-strike. Despite these differences, the COP errors we measure were consistent and within expectation with the errors in literature. For example, when predicting AP COP 50ms *before* foot-strike, we expect a RMSE<31.14mm (MAE<23.72mm). With this error, we might expect a torque error of 0.0954 Nm $kg^{-1}$ for fore-foot-first strikes, and 0.0895 Nm $kg^{-1}$ for rear-foot-first strikes, assuming a prediction error of 23.72 mm (Supplementary 1.1).

We also forecasted TOI prior to foot-strike at FHs between 16.67ms and 250.00ms. The TOI prediction error increased linearly with larger FH up to 166.67ms. Beyond this threshold, the error becomes non-monotonic as a result of the model beginning to default to constant values (fTOI constant values at the subject-level; min = 208.68ms, max = 218.60ms). The bias towards default or conservative predictions arises from the training paradigm. The loss was computed over the last 250ms before impact (each using a 250ms input context window). Despite the uniformly distributed ground truth TOI, the loss will tend to cause predictions to err towards the mean at the boundary (as the ground truth never lies outside of this boundary) to reduce overall error (*47–50*). This means that the model will tend to predict lower than the ground truth at longer FH, but higher than the ground truth at shorter FH. Paired with the fact that at longer FHs, there was less relevant visual information for the model to use, this conservative bias was more dramatic at longer, rather than, shorter FHs.

Similar to the COP forecasting task, the methods used to forecast TOI causally prior to foot-strike is novel. Even though the tasks are not directly comparable, the magnitude of error in the TOI prediction was similar to other published literature. For example, one study developed an algorithm to predict foot-strike time of impact during level ground treadmill running, based on 3D motion capture software, in post-processing (non-causal) for biomechanical analysis when force plates are not available. They found that most of their errors lied within +/- 20ms of the true foot-strike times based on the ground truth force plate data (*34*). A more recent study tackled this problem (again, using 3D motion capture software) with a machine learning-based solution and found their advanced algorithm presented most of its errors within +/- 10ms of the true foot-strikes (*35*). In comparison, our results demonstrate mean absolute errors (MAE) of <21.43ms for forecasts made <216.67 ms prior to foot-strike, using causal, vision-based data (*without* 3D markers).

As part of the study, we investigated how the model generalizes to dynamic variables in gait such as walking speed, which was explored in various biomechanical studies (*51–56*). We found that walking speed did not significantly affect COP or TOI error, suggesting that the model was robust to the range of walking speeds examined. However, ipsilateral toe velocity in the final swing phase (prior to the step) inversely affected the error in the COP task but was insignificant in the TOI task. In the COP case, this may have stemmed from the study's design. For instance, when subjects were instructed to target a specific foot landing (rear-foot, mid-foot, or fore-foot), they may have hesitated or accelerated during their final swing phase prior to the stair impact to control foot placement and achieve the target location, despite the researchers' best efforts to discourage such behaviour. Hesitations or accelerations may have elicited rapid changes in toe velocity, resulting in the swing phase being less natural and more difficult to model. One potential interpretation as to why the TOI error was agnostic to toe velocity could be due to differing mechanisms in which the TOI predictions were formed compared to the COP predictions. For instance, it could be hypothesized that the COP task was more difficult to model than the TOI task, since the COP prediction task needs to also encapsulate the impact time implicitly (to generate the predicted COP result upon impact). Further, changes in toe velocity (or hesitation in foot placement manifested through toe velocity changes), may result in changes to the precise position at which the foot impacts (COP), while affecting the actual impact time (TOI) less.

We also wanted to understand how the COP and TOI prediction errors generalize with various foot placement locations. While the TOI error was agnostic to differing foot placements, it was positively associated with the COP error. Specifically, more distal foot placements (i.e., higher COP) were associated with reduced predictive performance (higher error). One potential explanation for this higher error in fore-foot first landings was that small variations in foot orientation or movement may be amplified due to the longer lever arm and smaller contact area, making the COP more sensitive to small changes (e.g., in pitch or roll angles) and therefore be more difficult for the ML model to predict. In contrast, the foot placement did not affect the error in the TOI case significantly. The interpretation for this could be similar to that of the toe velocity case presented above. Namely, the TOI task was concerned with only the simpler problem of detecting impact time and hence, should be less dependent on the precise location of the impact. Finally, FH was a significant predictor in both the COP and (restricted linear domain) TOI tasks, as expected, since predictions formed earlier in time were physically further from the stair and therefore require longer extrapolation, which inherently involves greater uncertainty.

Since different subjects might exhibit different gait properties (*57–60*), we wanted to address subject-level variability. In general, inter-subject variability was more prominent in the COP task than the TOI task. More specifically, subject-specific intercepts (constant subject-level shifts in error) accounted for a similar amount of inter-subject variability as subject-specific forecast horizon slopes (with respect to error). The root of this inter-subject variability could be multi-faceted, ranging from individual differences in gait kinematics to ML model training dynamics, such as the ground truth structure (see Supplementary 1.2). Hyper-parameter tuning, increasing the dataset size, or increasing the model complexity may improve the system's ability to capture such patterns more consistently across subjects.

There were some notable areas at which experimental errors may have arisen during the collection and/or analysis. First, the instrumented insoles (Moticon Insoles, Moticon AG, Munich, Germany) may introduce measurement error in the derived COP, affecting the accuracy of the ground truth. One study found that the RMSE along the anterior-posterior axis was 15 +/- 6 mm for walking and 18 +/- 5 mm for running when comparing the instrumented insoles to gold standard force plate data (*61*). In the present study, force plates were infeasible as they would not fit on the stairs. Another source of measurement error could have arisen due to the discrete sampling of frames from the camera's frame rate. Namely, the impact frame was manually annotated through the inspection of frame-by-frame differences in each video. Given that the frames were discrete, there could be up to a +/- (maximum) 1 frame differential in terms of where the impacted frame was set to (i.e., if the impact was between frames). This could realize as irreducible error in the impact time of up to 16.67ms on a given measurement which could adversely affect both training and evaluation. A more precise method could be to estimate the impact time using a camera with a higher sampling rate. Third, the present study operated only on the right leg, resulting in the subject requiring to step with their right foot first on the stair, which may not represent real world scenarios or account for the effects of the individual's dominant foot. In this case, all but one subject was right-foot dominant. Finally, the act of instructing subjects to land with their fore-foot, midfoot, or rear-foot on the stair may not fully replicate real-world conditions, as it could alter their natural gait dynamics. However, this approach was required to capture a broad distribution of ground truth COP locations.

One other limitation of our findings was that our machine learning solution resembles a relatively narrow proof-of-concept rather than a fully adaptable solution. We only considered one set of stairs and defer generalization to other stairs or environmental features for future work. This may involve expanding the training dataset to incorporate variations of stairs or novel environmental features (i.e., curbs). Next, we trained subject-specific models rather than a generalizable subject-agnostic model in this case (see Materials and Methods for rationale). While it allows models to be tailored towards the specific intricacies of each subject, it likely will not generalize to novel subjects without subject-level calibration which will help the model learn subject-specific characteristics (i.e., fine-tuning the novel subject with 90 stair steps). Future efforts should prioritize generalization to new environments and subjects over mere performance optimizations.

Our findings confirm that both the COP and TOI tasks can be forecasted in real-time with reasonable accuracy using a lightweight vision-based model, even in low-data settings. Our subject-specific fine-tuning across only ~90 trials proved effective, and we quantified how factors such as FH, torso and toe velocity, and foot placement strategy influenced the prediction error. This proof-of-concept is a step towards more adaptable lower-limb assistive systems which mimic how the human visuo-locomotive system anticipates and responds to environmental changes in real-time, which can enable safer and more seamless mobility support. Despite this success, we recognize that expanding the subject sample size and performing other ML optimizations could further improve generalizability and performance. However, since this was a proof-of-concept, with the primary aim of assessing technical feasibility, these efforts were deferred to future work as they would likely yield diminishing insights relative to the resources required at this stage.

## MATERIALS AND METHODS
### Objectives and Design

The purpose of this experiment was to investigate the feasibility of estimating the impact COP and TOI during a gait transition from level-ground to stair-ascent using ML. The study was designed to collect data for the task across various subjects, specified walking speeds (slow = 1.0 m s$^{-1}$, medium = 1.25 m s$^{-1}$, fast = 1.5 m s$^{-1}$), and foot-strike locations (landing fore-foot, mid-foot, or rear-foot first) to be used to train causal ML models. The feasibility of the ML model operating in real-time was also investigated.

### Protocol

The following study protocol was approved by the University of Waterloo Research Ethics committee (University of Waterloo Ethics Review Board #45827). Eight individuals [n = 8, 2 females and 6 males; age = 22.8 ± 1.9 years (means ± SD); mass = 66.6 ± 5.7 kg; height = 172.5 ± 7.1 cm] who met the eligibility criteria were selected to participate. Subjects were 18-64 years old with no major issues pertaining to balance, muscular function, cardiac health, neural conditions, pain, or recent falls. Additional demographics are provided in Supplementary Table S1. Each subject was fitted with a camera (OAK-D wide, Luxonis, Denver, Colorado) on a 3D printed mount secured on the right shank with VELCRO® straps. We then instructed participants to place appropriately fitted sensorized insoles (Moticon Insoles, Moticon AG, Munich, Germany) in their shoes. Additionally, we fitted subjects with reflective markers on the top of their sternum (between the clavicles) and on the right foot over the second metatarsal head for use with a motion capture system (Vicon, Oxford, UK). To ensure a wide range of approach velocities across trials, we instructed subjects to walk at three different speed categories. Before beginning, we calibrated each subject's perception of walking speed categories (slow = 1.0 m s$^{-1}$, medium = 1.25 m s$^{-1}$, fast = 1.5 m s$^{-1}$) by monitoring the live differentiated sternum marker and providing feedback to help them associate each perceived speed with the corresponding walking velocity. During the trial, we instructed subjects to walk from level ground (i.e., take a few steps) and transition onto the first step of the stair such that their fore-foot, mid-foot, or rear-foot of the right foot would first contact the stair edge. We instructed participants to follow through up the remaining steps of the stairs to replicate natural stair climbing gait. For each collection, recordings from the motion capture system, camera, and insoles were initiated at the start of the trial. We targeted a total of 90 trials per subject, aiming to produce relatively wide distributions of each combination of speed and foot strike location. Ground truth COP distributions can be found in Supplementary Fig. S12.

### Data Acquisition and Processing
#### *Video*

A stereoscopic camera (OAK-D wide, Luxonis, Denver, Colorado) was mounted to a custom 3D printed bracket (made of TPU) secured to the shank with VELCRO® straps (Fig. 1A). The camera was positioned such that the entire foot, along with a substantial area of the terrain in front of the foot, was visible. The two stereo cameras captured videos at a resolution of 1280 by 720 at 60 frames per second (FPS) with 150°/128°/80° diagonal, horizontal, and vertical field of views (FOV). Furthermore, the camera interfaced with a laptop for processing via a USB 3.0 SuperSpeed cable and data was subsequently collected and stored through a custom Python script. Left and right video channels were first down sampled (bilinear interpolation) to a resolution of 50 by 25 resulting in a 2:1 aspect ratio. For each trial, we determined the impact frame index (the first frame at which the foot contacts the stair) through manual annotation using custom software. Further, all pixel intensities for a given frame were min-max normalized. Once processed, videos were stored as PyTorch tensors (see Supplementary 1.3 for more details).

***Sensorized Insoles***

Sensorized Insoles (Moticon Insoles, Moticon AG, Munich, Germany) were used to determine the center of pressure upon foot strike on the first step of the stairs from a level ground transition. The insoles measured the AP center of pressure (COP) at 100 Hz as a normalized (by length) quantity bounded on the inclusive range of [-0.5, 0.5]. This quantity was not further scaled prior to machine learning training. For a given transition onto the stair, ground truth was extracted upon foot-strike on the stair automatically with a custom algorithm. We manually vetted the results for precision.

***Motion Capture***

A motion capture system (Vicon, Oxford, UK) was used to record the position of markers on each subject as they move in 3D space at 200 Hz. In this study, the torso and toe markers were specifically used to determine the approach and toe velocity respectively. In each trial, the environment was setup such that the subject approached the stairs along the x-axis, ascended along the z-axis, with little to no lateral movement along the y axis. After data collection, we performed relevant post-processing, including gap filling and exporting marker trajectory and velocities to comma separated values (csv). In cases where Vicon trials were corrupted, we discarded the entire trial from the subsequent analysis. Further, we developed custom algorithms to extract the torso and toe velocities per trial (see Supplementary 1.4, 1.5).

**Machine Learning Model**
***Architecture***

We trained separate machine learning models to forecast COP and TOI quantities from the video data. In both tasks, we used a serially arranged CNN (feature extractor) to RNN (sequence regressor) architecture. The goal of the CNN was to take input from a pair of stereoscopic frames at each timestep and extract an abstract representation of relevant spatial information. For instance, an extracted feature vector might contain information on the relative position of the foot to the stair edge. Then, we used an RNN to process these features temporally, receiving both the extracted CNN features at a given timestep and a vector of hidden states as inputs. These hidden states contain implicit information regarding the spatiotemporal dynamics from previous timesteps. At each timestep the RNN outputs a scalar value, serving as the prediction for that particular timestep. This RNN, which was unidirectional, which was vital to ensure the model was causal since future frames will contain information on the impact itself.

The CNN comprised three layers, each containing three 3x3 convolutional modules (stride=1) with ReLU6 activation functions, followed by a 2x2 max pooling module (stride=2). Inputs to each convolutional layer exhibited "same" padding. Each layer approximately reduced the spatial resolution by a factor of 2. Additionally, feature map depth was expanded for the first two blocks, first increasing the number of channels from 2→12 and later increasing from 12→32. The third layer maintains the 32-channel depth. Note that the input tensor has two channels for the use of left and right stereo substituents of the camera.

The RNN contained two serially arranged RNN blocks with hyperbolic tangent nonlinearity. The first RNN block accepted the flattened features extracted by the CNN, and its output was passed to the second RNN block. Both RNNs maintained their own hidden state across timesteps. The output from the second RNN was the scalar corresponding to the predicted COP or TOI at that timestamp. The CNN and RNN contained 47, 437 and 34, 147 trainable parameters respectively, totaling to 81, 584 in the combined model. Logically, the CNN and RNN exhibit inductive biases that assist the task at hand, which further described in Supplementary 1.6.

### Model Fine-Tuning

To expediate machine learning training, we fine-tuned base models trained on a separate but related level-ground to stair-stepping task (see Supplementary 1.7). To estimate model performance on each subjects' dataset individually (~n=90, Supplementary Table S1), we performed leave-one-out-cross-validation (LOOCV). When performing this fine-tuning, we used the same hyper-parameters that were used the train the base model, except for the learning rate, number of training epochs, and the input context window. We reduced the learning rate by a factor of 10, which is common practice for fine-tuning models (*62*, *63*). Further, and to eliminate risk of over-fitting or optimal hyper-parameter leakage, we pre-set the number of training epochs to 250 (post-hoc visualization confirmed that this was a good choice) (*64*). Finally, since the *base model* was trained statically across 100 frames (See supplementary 1.7), forecasts at different times contained varying input context (i.e., predictions at 15 frames before impact had 85 frames of input, but predictions at 5 frames before impact had 95 frames of input). In contrast, during fine-tuning, each forecast at any moment relied on data from the preceding 15 (250ms) camera frames, hence providing a constant input context window for each forecast. Through choosing hyper-parameters a priori to training, we were able to estimate model performance in an unbiased and computationally feasible way. Other methods such as nestled cross-validation with hyper-parameter tuning could theoretically improve results, however, was deemed computationally infeasible in this setting.

This simulates a real-world scenario where per-subject calibration is feasible (only requiring ~90 steps) and potentially beneficial. This approach allows for the model to be tailored towards individualized patterns and dynamics, which may be averaged out in pooled training, unless the training data contained a large amount of subject diversity. Furthermore, even with diverse data, a simplistic and compact model (proposed here) is unlikely to capture these across-subject dynamics. Hence in this case, there may be a need to increase the model complexity, which will inherently reduce the real-time capability of the model. Rather, using LOOCV over each subject individually allows the model to be tailored towards the nuances of each subject, while permitting less computationally expensive models, at the cost of subject-level calibration.

### COP Fine-Tuning

To fine-tune the COP task, we computed loss as mean-squared-error (MSE) between the model output and the normalized ground truth COP over the last 250ms preceding stair impact. This 250ms window was chosen for relevance and computational efficiency. Notably, since the last 250ms would correspond to the latter half of the swing phase in an average gait cycle (*65*), it was a meaningful window for loss computation.

### TOI Fine-Tuning

Fine-tuning the TOI task was analogous to the COP task. In this case, the ground truth was computed through back counting the frames prior to the annotated impact frame. For instance, the last 4 frames leading up to foot-strike for a given video sample would correspond to ground truths of [1, 2, 3, 4] or equivalently [1/60, 2/60, 3/60, 4/60] seconds. In the case presented here, ground truths varied from [1, 15] frames or [0.0167, 0.250] seconds, for the same reasons as presented in the COP case.

## Analysis
### Mean Absolute Error at each Forecast Horizon
To quantify the error at each FH, we computed the mean-absolute-error (MAE) between the predicted quantity (either COP or TOI) and their associated ground truth. This was computed over the unnormalized quantities, where COP was rescaled based on insole length in millimetres (0 mm is indicative of the rear-foot), and TOI was rescaled to milliseconds prior to impact based on the camera's frame-rate (60 FPS – 1/60s period). We chose to use MAE as the reporting metric due to its interpretability benefits, as well as being on the same units and scale as the original quantity. For example, in the COP case, error can be directly interpreted as "on average, the COP prediction was off by x mm". While the RMSE (square root of MSE) also reports errors in the same units as the original quantity, it was less easily directly interpreted since it penalizes larger errors disproportionately compared to smaller errors due to its quadratic nature. Hence, we use the MAE to report error at each FH to give a sense of how the average error changes as predictions were formed earlier in time prior to foot-strike. The MAE provides a summary statistic over the error between the forecasted value and ground truth value. The fCOP vs COP scatter plots were also visualized (for select FH and subjects) to observe the distribution of the error.

### Residual Analysis
For the COP task, we plotted residuals on a scatter plot against ground truth. This allowed for visualization towards how the residuals were spread as a function of foot location. To quantify model systematic (conditional) bias, we fit ordinary-least-square linear regressions on these plots. Here, the slopes indicated a tendency towards over- or under-prediction based on the foot location (i.e., systematic bias at the foot level). Additionally, we visualized the residuals at each FH (and between subjects) using boxplots for the COP and TOI tasks. This allowed for an assessment of systematic bias at the FH level (i.e., through the median of the box-plots), as well as a visual assessment of the variance of the residual distributions. For the TOI task, systematic bias was also observed through plotting the mean fTOI for a given FH versus the ground truth TOI, for each subject.

### Mixed Effects Modelling
To quantify the effect of various predictors and evaluate subject-level variation in the prediction task, we conducted linear mixed effects modelling for the COP and TOI tasks independently. Four fundamental predictors were standardized and used in this analysis: (1) FH, (2) approach (torso) velocity, (3) toe velocity, and (4) ground truth COP. In both tasks, it was assumed that the effect of the approach velocity, toe velocity, and ground truth COP on the errors would be non-subject dependent (fixed effects) which helped to simplify the model. Furthermore, we introduced random effects at the subject-level by using random intercepts and random slopes with respect to FH. Including subject-grouped random intercepts acknowledges and captures the concept that some individuals may have systematically higher or lower errors, irrespective of the predictors. Further, including subject-grouped random slopes allowed us to understand how the error might increase at variable rates across subjects when predictions were formed further into the future. Both of these random effects were motivated from the MAE-FH curves (Supplementary

Fig. S2, S7). In both the COP and TOI tasks, the absolute values of the errors were used rather than their signed counterparts since interest lied within the magnitude of the error in this analysis, not the direction. Further, signed errors may cancel out for unbiased errors (i.e., error with respect to FH in the COP model was mainly due to the increase in variance), which could skew the perception of errors. Before fitting the initial mixed-effects model, we verified key assumptions, including testing homoscedasticity and normality of the residuals, and checking the linearity between the predictors variables and the response variable, to ensure the robustness of our results (see Supplementary 1.8 for more details on the verification of these assumptions). Our analysis revealed heteroscedasticity in the linear mixed effects model's residuals. To resolve this, we used a cube root power transformation on the response variable (*66, 67*). In our analysis, statistically insignificant fixed effects ($p>0.05$) were removed after the initial fitting, and the model was subsequently re-fit to ensure predictor independence for interpretability (i.e., if insignificant effects were correlated to significant effects). In the final model, we assessed the practical significance of the un-standardized and back-transformed coefficients for statistically significant effects ($p<0.05$) by rescaling to a more interpretable range. The derivation for the approximately unbiased back-transformation (to handle Jensen's inequality) can be found in Supplementary 1.9). Reported results were back-transformed unless explicitly stated. All analysis was conducted in R with the lmerTest package.

**Live Simulation**

To estimate the real-time capabilities of the proposed models, we conducted a simple frame-rate test. We setup a pipeline that ran the COP and TOI models live when interfacing with the camera. Since RNN-based methods maintain states through time, the hidden states were appropriately propagated across timesteps (frames). For each trial of the test, we first loaded in model weights from a random LOOCV trial. Then, we instructed a subject to perform the level ground to stair ascent stepping task (that was a done during original data collection) continuously over a duration of approximately 2 minutes. Further, we computed the effective frame rate as the number of frames processed (i.e., predictions formed on) divided by the total elapsed time when running live. This was repeated three times in order to assess variability. While this does not account for all real-time constraints, it provides a rough estimate of real-time feasibility. We conducted this test on a laptop (MacBook Air, Apple M2 chip, 16GB unified memory) and an edge device (Jetson Orin Nano, 8 GB LPDDR5, 6-core ARM Cortex-A78AE CPU, 1024-core Ampere GPU).

## Acknowledgments

### General

The authors would like to thank Ethan Luk for his role in assisting in data collections and in the development of the torso velocity extraction code. The authors would also like to thank Alyson Colpitts for her help with the motion capture system.

### Funding:

This work has been supported though University of Massachusetts Lowell (RWN), University of Waterloo (RWN), Natural Sciences and Engineering Research Council of Canada (NSERC) RGPIN-03728-2023 (RWN) and RGPIN-2022-03878 (JT).

### Author contributions

Conceptualization: MDM, RWN, JT
Methodology: MDM, RWN, JT
Investigation: MDM, RWN
Visualization: MDM, RWN, JT
Funding acquisition: JT, RWN
Project administration: RWN
Supervision: RWN, JT
Data Curation: MDM
Software: MDM
Writing – original draft: MDM
Writing – review & editing: MDM, RWN, JT

### Competing Interests

The authors declare that they have no competing interests.

### Data and materials availability:

Data and designs will be available upon reasonable researcher request (RWN, richard_nuckols@uml.edu).